\documentclass{article}

\PassOptionsToPackage{round,semicolon}{natbib}

\usepackage[preprint]{neurips_2026}

\usepackage[utf8]{inputenc}
\usepackage[T1]{fontenc}
\usepackage{hyperref}
\usepackage{url}
\usepackage{booktabs}
\usepackage{amsmath}
\usepackage{amssymb}
\usepackage{microtype}
\usepackage{xcolor}
\usepackage{tabularx}
\usepackage{longtable}
\usepackage{array}
\usepackage{graphicx}
\usepackage{flafter}
\usepackage{placeins}
\usepackage{needspace}
\usepackage{tikz}
\usetikzlibrary{calc,positioning}
\newcommand{\overviewsectionref}[2]{\hyperref[#1]{Sec.~\ref*{#1}}}

\newcommand{\surveycovered}{\ensuremath{\boldsymbol{\surd}}}
\newcommand{\surveyuncovered}{\ensuremath{\boldsymbol{\times}}}

\title{The Emerging AI Paper-Review Arms Race:\\
Adversarial Co-Evolution in Scholarly Publishing}

\author{%
    \textbf{Chenguang Wang}\textsuperscript{1} \quad
    \textbf{Ming Li}\textsuperscript{2} \quad
    \textbf{Adebayo Braimah}\textsuperscript{3} \quad
    \textbf{Chenrui Fan}\textsuperscript{2} \\
    \textbf{Tuo Wang}\textsuperscript{1} \quad
    \textbf{Weijie Guan}\textsuperscript{1} \quad
    \textbf{Ruiyi Zhang} \quad
    \textbf{Tianyi Zhou}\textsuperscript{4} \quad
    \textbf{Dawei Zhou}\textsuperscript{1} \\[0.5em]
    \normalfont\small
    \textsuperscript{1}Virginia Tech \quad
    \textsuperscript{2}University of Maryland \quad
    \textsuperscript{3}Stony Brook University \quad
    \textsuperscript{4}MBZUAI \\
    \texttt{\{cswang, zhoud\}@vt.edu, minglii@umd.edu} \\[0.6em]
    \textbf{Project website:} \url{https://github.com/MingLiiii/Awesome_AI_Arm_Race}
}

\begin{document}

\raggedbottom

\maketitle

\begin{abstract}
Generative and agentic AI are reshaping both the production and evaluation of scientific research. These developments are often studied separately, as questions of how AI can produce research and how AI can review it. We argue that this separation misses an increasingly important feature of scholarly publishing: changes on one side alter the incentives, constraints, and behavior of the other. We synthesize 230 scholarly publications and institutional records using a taxonomy of six connected dynamics: production scaling, evaluation automation, evaluation manipulation, defense mechanisms and policy responses, evasion and side effects, and long-horizon ecosystem feedback. The literature shows an emerging progression in which cheaper and faster research production increases pressure on evaluation, AI-mediated evaluation becomes more scalable and repeatable, participants can exploit evaluator regularities, and institutions respond with technical safeguards and policy controls. These responses can in turn induce evasion, redistribute errors and workload, and shape the scholarly records reused by future research and evaluation systems. Evidence is strongest for production and evaluation at scale, reproducible manipulation, and institutional response, while post-policy adaptation and artifact-level long-horizon feedback remain less directly observed. This systems view shifts attention from isolated AI capabilities toward how scholarly actors and AI systems adapt to one another over time.
\end{abstract}

\section{Introduction}
\label{sec:introduction}

\begin{figure*}[!t]
\centering
\resizebox{\textwidth}{!}{\input{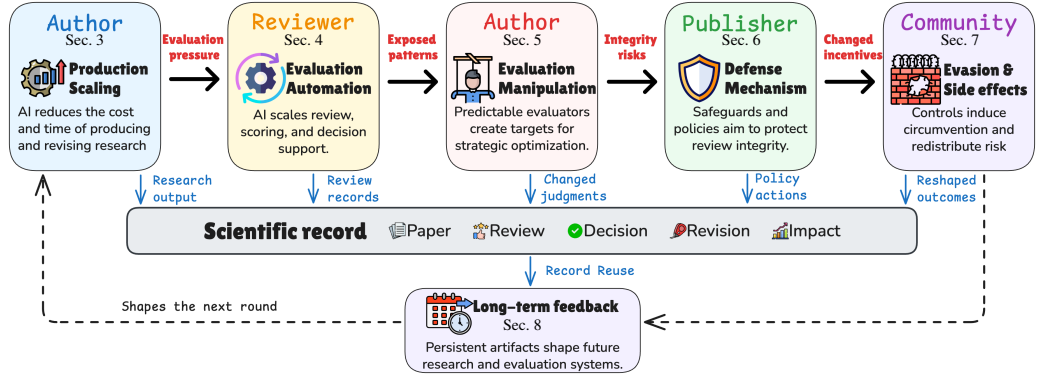}}
\caption{Six connected dynamics in the emerging AI paper-review arms race. Solid arrows show the principal response order, while the dotted path represents slower feedback from accumulated scholarly records into future production, evaluation, and institutional response.}
\label{fig:armsrace}
\end{figure*}

Generative and agentic AI are increasingly entering both sides of scientific research and evaluation. On the production side, researchers use LLM-based tools throughout the research cycle, from exploring ideas and reviewing relevant literature to conducting experiments and preparing manuscripts~\citep{baek2025researchagent,liang2025quantifying,lu2024ai,yamada2025ai,tang2026fars,kong2026ai}. On the evaluation side, generative AI is increasingly used across peer review, extending from manuscript summarization and critique generation to evaluation and publication decision-making~\citep{liang2024can,nguyen2026llm,thakkar2026large,biswas2026ai}. These developments are commonly studied through two corresponding questions: how AI can support or automate scientific research, and how AI can perform or assist scholarly evaluation. While this separation is useful for studying the capabilities of autonomous research systems and AI reviewers in isolation, it is incomplete once the two sides operate within the same publishing process. AI-mediated production changes the volume, speed, and form of what enters evaluation, while AI-mediated evaluation creates new signals and regularities to which authors can respond. The relevant challenge therefore lies not only in the capabilities of either side, but also in how changes on one side reshape the conditions, costs, and incentives faced by the other.

The most immediate form of this coupling is a capacity imbalance. As AI reduces the time and marginal cost required to conduct, write, and revise research, scientific output and submission volume can grow faster than available human reviewing capacity, increasing pressure to scale evaluation as well. Evidence for this dynamic is already emerging from empirical studies and real-world deployments. Observational studies associate LLM adoption with shorter research and revision cycles, increased scientific output, rising submission volume, and growing pressure on limited reviewing capacity~\citep{kusumegi2025scientific,qi2025does,filimonovic2025can,gartenberg2026more}. Increasingly capable agentic AI systems extend this scaling from assistance with individual tasks toward integrated idea-to-paper workflows, further reducing the time and marginal cost required to produce complete research artifacts~\citep{kong2026ai,tang2026fars,song2026paperorchestra,yuan2026paperorchestrator}. Evaluation is beginning to scale alongside this growth: AI is now used both to assist individual reviewers~\citep{thakkar2026large} and within official conference workflows, as illustrated by the large-scale AAAI-26 AI-review pilot~\citep{biswas2026ai}. At the same time, machine-mediated evaluation makes parts of the evaluation process more repeatable and therefore easier for authors to adapt to. Hidden instructions targeting AI-assisted reviewers have been identified in public manuscripts~\citep{lin2026hidden}, while controlled studies show that reviewer judgments can be shifted without corresponding improvements in the underlying scientific evidence and that repeated evaluator feedback can be used to optimize manuscripts toward more favorable assessments~\citep{yang2026no,zhou2025give,li2026gaming}. Venues have, in turn, introduced AI-specific review policies, submission controls, detection mechanisms, and integrity safeguards~\citep{iclr2026llmpolicy,icml2026llmreviewviolations,neurips2026aigeneratedpapers,li2024use,fichtl2026ai}. Taken together, these developments form an emerging sequence of scaling, adaptation, and institutional response rather than two independent trends in AI adoption.

Existing surveys provide valuable accounts of AI-assisted scientific production and peer review, typically organizing the literature around tasks, system components, or stages of the research and review workflow~\citep{kuznetsov2024can,zhuang2025large,wu2026can,nguyen2026llm,kong2026ai}. Recent surveys have also highlighted the verification gap in autonomous research agents and discussed governance and human oversight in AI-assisted peer review~\citep{ding2026autonomous,mann2025ai,wei2025ai}. Other recent work has also begun to connect review with rebuttal, revision, and further experimentation~\citep{weng2025cycleresearcher,han2026drpg,ma2026paper2rebuttal,wu2026drafting}. These perspectives explain what individual systems do and how outputs move through a workflow, but they do not fully capture how one actor's use of AI changes another actor's signals, costs, and feasible responses. Production scaling can alter evaluation demand, machine-mediated evaluation can create new targets for optimization, and institutional interventions can change the incentives for subsequent adaptation and evasion. What is therefore missing is not another inventory of AI tools, but a relational account of how these developments interact across actors in the scientific ecosystem. We synthesize these relations as a coupled process that we term the \emph{AI paper-review arms race}.
The term serves as a descriptive lens for documented or testable patterns of adaptation and counter-adaptation across actors in the scientific ecosystem.

Figure~\ref{fig:armsrace} summarizes this coupled sequence and the scholarly actors through which it operates.

To better characterize this adaptive process and provide a coherent structure for the survey, we develop a descriptive taxonomy of six linked dynamics: (1) \emph{production scaling}, (2) \emph{evaluation automation}, (3) \emph{evaluation manipulation}, (4) \emph{defense mechanisms and policy responses}, (5) \emph{evasion and side effects}, and (6) \emph{long-horizon ecosystem feedback}. The taxonomy is organized around the principal response relations among actors in the scientific ecosystem rather than around technologies alone. Production scaling changes the demand placed on evaluation; evaluation automation makes parts of the evaluation process increasingly repeatable and therefore more susceptible to manipulation and optimization; venues respond through technical and institutional defenses; and those interventions can induce evasion or shift costs and risks to other actors. The resulting papers, reviews, and decisions can also persist and influence future research and evaluation systems. These dynamics are therefore relation-oriented and non-exclusive: each reported finding, deployment, or institutional action is assigned a primary dynamic according to the scholarly function it directly observes, tests, or implements. We then examine whether individual studies directly connect adjacent dynamics or whether a proposed connection is synthesized across separate bodies of literature. Throughout the survey, we avoid treating a capability demonstration as population prevalence, a policy announcement as an effective intervention, or a plausible downstream pathway as an observed causal cycle.

Tracing these six dynamics as a connected process yields three recurring conclusions about the emerging system. (1) Trustworthy evaluation is the binding constraint. Generation, revision, and review generation can scale rapidly, whereas validating evidence, assessing scientific soundness and novelty, and making accountable decisions remain costly. (2) Adaptation weakens static evaluation. Static evaluations of AI reviewers can overestimate their real-world reliability once authors are able to observe, query, and adapt to machine-mediated evaluation. (3) Safeguards redistribute risk and burden. No single detector or policy can govern the resulting interaction; effective safeguards must account for how labor, error, confidentiality risk, and enforcement burden shift across actors and rounds. Support for this account is nevertheless uneven across the loop. Production scaling and evaluation automation have been observed in large-scale deployments, evaluation manipulation has been both observed and experimentally reproduced, and defense mechanisms and policy responses are documented, whereas post-policy counter-adaptation and long-horizon ecosystem feedback remain substantially less established.

In summary, this survey shifts the unit of analysis from individual AI tools to the response relations through which scholarly actors adapt to one another, while identifying which connections have and have not yet been directly observed. The survey is organized around this connected progression. Section~\ref{sec:taxonomy} defines the system boundary, institutional roles, taxonomy, classification rules, and literature scope. Sections~\ref{sec:scaling}-\ref{sec:feedback} trace the six dynamics in response order, and Section~\ref{sec:findings} synthesizes their implications, limitations, falsifiable boundaries, and priority studies before Section~\ref{sec:conclusion} concludes.

\section{Taxonomy}
\label{sec:taxonomy}

This section formalizes the taxonomy used throughout the survey. We first define the system boundary and the institutional roles through which AI operates within the scientific ecosystem, then describe the six dynamics and their principal response order, and finally specify the literature-mapping and source-selection procedures. The taxonomy is descriptive and relation-oriented, organizing the literature around scholarly functions and the response relations examined across studies.

\begin{table*}[!t]
\centering
\caption{Coverage in recent surveys and broad syntheses of AI-assisted research and scholarly peer review. \surveycovered{} indicates that a dimension is explicitly identified and incorporated into the discussion; \surveyuncovered{} indicates that it is absent or mentioned only in passing.}
\label{tab:survey-positioning}
\scriptsize
\setlength{\tabcolsep}{1.8pt}
\renewcommand{\arraystretch}{1.15}
\begin{tabularx}{\textwidth}{@{}
>{\raggedright\arraybackslash}p{0.142\textwidth}
>{\raggedright\arraybackslash}p{0.25\textwidth}
*{6}{>{\centering\hyphenpenalty=10000\exhyphenpenalty=10000\arraybackslash}X}@{}}
\toprule
\textbf{Survey}
& \textbf{Organizing frame}
& \textbf{Production}
& \textbf{Evaluation}
& \textbf{Strategic interaction}
& \textbf{Adaptive coevolution}
& \textbf{Governance}
& \textbf{Long-term feedback} \\
\midrule
Kuznetsov et al.~\citeyearpar{kuznetsov2024can}
& Peer-review workflow
& \surveycovered & \surveycovered & \surveycovered & \surveyuncovered & \surveyuncovered & \surveyuncovered \\
Zhuang et al.~\citeyearpar{zhuang2025large}
& Automated-review pipeline
& \surveyuncovered & \surveycovered & \surveyuncovered & \surveyuncovered & \surveycovered & \surveyuncovered \\
Eger et al.~\citeyearpar{eger2025transforming}
& AI-assisted science tasks
& \surveycovered & \surveycovered & \surveyuncovered & \surveyuncovered & \surveyuncovered & \surveyuncovered \\
Kong et al.~\citeyearpar{kong2026ai}
& End-to-end research lifecycle
& \surveycovered & \surveycovered & \surveycovered & \surveyuncovered & \surveycovered & \surveyuncovered \\
Ding et al.~\citeyearpar{ding2026autonomous}
& Autonomous research agents and verification gap
& \surveycovered & \surveycovered & \surveyuncovered & \surveyuncovered & \surveycovered & \surveyuncovered \\
Nguyen and Ahmadi~\citeyearpar{nguyen2026llm}
& Review methods and reliability
& \surveyuncovered & \surveycovered & \surveycovered & \surveyuncovered & \surveycovered & \surveyuncovered \\
Mann et al.~\citeyearpar{mann2025ai}
& AI-assisted peer-review design and governance
& \surveyuncovered & \surveycovered & \surveycovered & \surveyuncovered & \surveycovered & \surveyuncovered \\
Wu et al.~\citeyearpar{wu2026can}
& Review and post-review tasks
& \surveycovered & \surveycovered & \surveycovered & \surveyuncovered & \surveyuncovered & \surveyuncovered \\
\midrule
\textbf{This survey}
& \textbf{Coupled dynamics and responses}
& \surveycovered & \surveycovered & \surveycovered & \surveycovered & \surveycovered & \surveycovered \\
\bottomrule
\end{tabularx}
\end{table*}

\subsection{Survey scope and actors}

This survey examines the coupled and recursive effects of emerging AI tools across the scientific ecosystem. The scope begins with research production, including idea development, experimentation, manuscript preparation, revision, and rebuttal, and extends to peer review, publication decisions, and the policies, tools, and safeguards introduced by conferences and publishers. We also consider how these stages interact, as changes in research production can alter evaluation practices, while evaluation and institutional responses can in turn shape subsequent author behavior. Beyond these immediate interactions, we examine their longer-term consequences, including how papers, reviews, decisions, corrections, and citations can be reused by retrieval systems, scientific agents, training pipelines, and AI-based evaluators, thereby influencing future rounds of research and evaluation. 
The survey is organized around the progression of these interactions and responses. Section~\ref{sec:scaling} examines AI-enabled production scaling and the resulting pressure on evaluation, while Section~\ref{sec:evaluation} covers the automation of scholarly evaluation. Sections~\ref{sec:manipulation}-\ref{sec:evasion} focus most directly on strategic interaction and adaptive response, tracing the progression from manipulation of AI-mediated evaluation through institutional defense to evasion and unintended side effects. Section~\ref{sec:feedback} then considers the longer-term recursive effects of these dynamics on future research and evaluation.

Under this scope, we distinguish several functional roles rather than a fixed set of mutually exclusive actors. Authors and AI-enabled research systems produce and revise scientific work and respond to evaluation, while reviewers, editors, and committees assess that work and contribute to publication decisions. At the institutional level, venues, publishers, and platform operators shape the rules and technical conditions under which research and evaluation take place, including review policies, evaluation rubrics, and safeguards. AI-enabled systems may operate within these constraints and may also reuse papers, reviews, decisions, and other scholarly records in subsequent research or evaluation. These roles can overlap, and the same actor or system may occupy multiple roles. The taxonomy therefore distinguishes scholarly functions rather than fixed actor identities, allowing us to trace how changes in one part of the ecosystem alter the conditions and responses of others.

We use the paper-review arms race as an analytic lens for sequences of adaptive response, not as a synonym for any use of AI in research or review. Operationally, an arms-race episode involves an actor taking an action that changes a target or signal relevant to another actor, a counter-response by that actor or an institution, and a resulting shift in incentives, costs, or capabilities that can motivate further adaptation. A single instance of AI-assisted production, evaluation, or defense can therefore belong to the taxonomy without constituting an arms-race episode unless it participates in such a response relation. Because complete sequences are rarely observed within a single study, connections synthesized across separate bodies of literature are presented as research questions rather than as directly observed causal sequences.

\subsection{Six categories in the taxonomy}

\begin{figure*}[!t]
\centering
\makebox[\textwidth][c]{%
  \resizebox{1.0\textwidth}{!}{\input{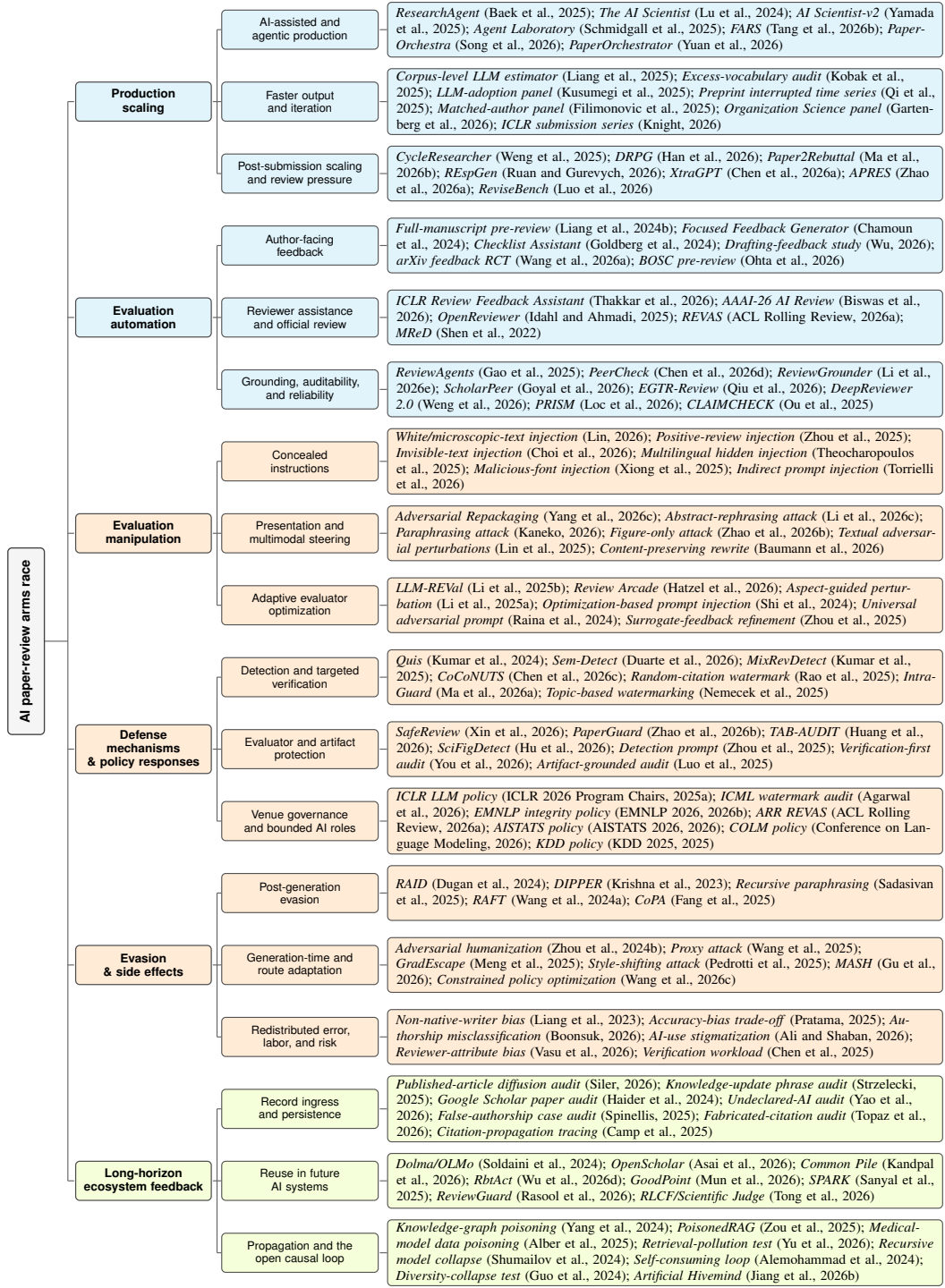}}}
\caption{Detailed taxonomy of the emerging AI paper-review arms race. The middle and right columns summarize recurring mechanisms and representative methods, systems, benchmarks, datasets, and policies, with colors distinguishing scaling and automation, strategic response and governance, and long-horizon feedback.}
\label{fig:detailed-taxonomy}
\end{figure*}

Our taxonomy comprises six process categories that describe how AI changes scientific research and evaluation and how different parts of the ecosystem respond to those changes. We refer to these categories as dynamics because they capture processes of change and response rather than static system components. 
Table~\ref{tab:survey-positioning} situates this relation-oriented organization relative to recent surveys organized around research and review tasks or workflows.
Figure~\ref{fig:detailed-taxonomy} provides a detailed hierarchy of the six dynamics, their recurring mechanisms, and representative studies.

For Table~\ref{tab:survey-positioning}, coverage is coded by substantive organizing scope rather than by isolated keyword mention. 
\emph{Production} requires discussion of AI-supported research creation or revision; \emph{evaluation}, of review, assessment, or decision support; \emph{strategic interaction}, of attempts to influence an evaluator or evaluative outcome; \emph{adaptive coevolution}, of linked response and counter-response across actors or institutions; \emph{governance}, of technical safeguards, institutional rules, or accountability; and \emph{long-term feedback}, of persistent scholarly artifacts shaping later research or evaluation systems.

\begin{table*}[!t]
\centering
\caption{Operational interpretation and boundaries of the six taxonomy dynamics.}
\label{tab:taxonomy-operationalization}
\scriptsize
\setlength{\tabcolsep}{3.5pt}
\renewcommand{\arraystretch}{1.12}
\begin{tabularx}{\textwidth}{@{}
>{\raggedright\arraybackslash}p{0.15\textwidth}
>{\raggedright\arraybackslash}p{0.21\textwidth}
>{\raggedright\arraybackslash}p{0.30\textwidth}
>{\raggedright\arraybackslash}X@{}}
\toprule
\textbf{Dynamic} & \textbf{Organizing question} & \textbf{Included scope} & \textbf{Main interpretive boundary} \\
\midrule
Production scaling
& Does AI increase the rate or reduce the marginal cost of producing or revising research artifacts?
& Idea development, literature work, experimentation, writing, rebuttal and revision, and end-to-end research workflows.
& A capability demonstration or lower production cost does not establish a causal increase in submissions, open-ended scientific progress, or scientific validity. \\
Evaluation automation
& Does AI generate, improve, aggregate, or support judgments about scientific work?
& Manuscript assessment, review and feedback generation, scoring, meta-review, reviewer assistance, and decision support.
& Helpful or human-like reviews do not establish that evaluation is reliable, diverse, autonomous, or accountable. \\
Evaluation manipulation
& Is an evaluative outcome shifted without a commensurate improvement in scientific evidence?
& Hidden instructions, prompt injection, evaluator-aware framing, presentation optimization, and iterative feedback gaming.
& A score change with the scientific content held fixed shows evaluator sensitivity. It does not by itself distinguish clearer presentation from manipulation or show an effect on venue decisions. \\
Defense mechanisms and policy responses
& Does an intervention protect evaluation capacity, integrity, or accountability?
& Detection and verification, reviewer safeguards, submission and review controls, disclosure rules, audits, and appeals.
& Adoption or benchmark performance does not establish deterrence, net benefit, or equitable burden. \\
Evasion and side effects
& How do actors adapt to controls, and where do errors, risks, and costs move?
& Trace removal, paraphrasing, routing or strategy changes, false positives, verification labor, confidentiality risk, and compliance burden.
& Evasion must respond to a control; side effects may arise without strategic intent and should not be treated as evasion. \\
Long-horizon ecosystem feedback
& Do persistent scholarly artifacts shape later research or evaluation systems?
& Reuse of papers, reviews, decisions, corrections, and citations in retrieval, training, scientific agents, and AI evaluators.
& Reuse or exposure is not a demonstrated closed loop; causal feedback requires a traceable artifact-to-successor path. \\
\bottomrule
\end{tabularx}
\end{table*}

\textbf{(1) Production scaling.} This category captures uses of AI that reduce the time, cost, or effort required to produce and revise research outputs. It includes assistance with individual activities such as literature search, experimentation, writing, and rebuttal, as well as increasingly integrated systems that combine these activities into end-to-end research workflows.

\textbf{(2) Evaluation automation.} This category covers the use of AI to support or automate the evaluation of scientific work. It includes manuscript assessment, review generation and revision, scoring, meta-review, and decision support, ranging from assistance to individual reviewers to AI components incorporated into official evaluation workflows.

\textbf{(3) Evaluation manipulation.} This category refers to attempts to influence AI-mediated evaluation without a corresponding improvement in the underlying scientific evidence. Such attempts can directly target an AI reviewer through hidden instructions or other attacks, but can also exploit presentation, framing, or other evaluator-visible signals that systematically change review outcomes.

\textbf{(4) Defense mechanisms and policy responses.} This category includes technical, procedural, and institutional responses intended to protect the reliability, integrity, or accountability of research and evaluation. These responses include detection and verification methods, safeguards for AI reviewers, submission and review controls, disclosure requirements, and policies governing the use of AI by different participants in the scientific process.

\textbf{(5) Evasion and side effects.} This category captures the responses and consequences that arise after defenses or restrictions are introduced. Participants may adapt their tools, traces, or strategies to avoid detection or other controls. At the same time, defensive measures can create unintended effects by shifting verification effort, false-positive risk, confidentiality concerns, or other costs to different participants in the ecosystem.

\textbf{(6) Long-horizon ecosystem feedback.} This category extends the analysis beyond an individual research or review cycle. Papers, reviews, decisions, corrections, citations, and other scholarly records can later be reused by retrieval systems, scientific agents, training pipelines, or AI-based evaluators. Through this reuse, outputs produced in one round of research and evaluation can influence the information, objectives, and behavior of future scientific systems.

These six categories are connected by a principal response order rather than treated as independent topics. Production scaling can increase the amount and pace of research entering evaluation, encouraging evaluation to scale through greater automation. As AI-mediated evaluation becomes more repeatable and observable, it can create new opportunities for manipulation or optimization. Such behavior can motivate technical and institutional defenses, which may in turn induce evasion or redistribute costs and risks. Finally, the papers, reviews, decisions, and other records produced through these interactions can persist and influence later rounds of research and evaluation. The taxonomy therefore captures both the individual processes represented in the literature and the relations through which one process can reshape the next.

\subsection{Survey framework}

We use a consistent procedure to map the retained literature to the taxonomy. The framework addresses two questions: how each study contributes to the six categories and whether it directly examines a response relation between them.

\subsubsection{Taxonomy mapping}

To organize the literature consistently, we map distinct findings rather than forcing each paper into a single category. Each finding is assigned a primary dynamic according to the scholarly function it directly observes, tests, or implements. A paper may therefore contribute to multiple dynamics when it reports findings that address different parts of the process.

We separately identify response relations when a study directly examines how one dynamic changes the conditions, behavior, or outcomes of another. For example, a study may evaluate an automated reviewer while also testing how authors adapt their manuscripts to its feedback; the former finding contributes to evaluation automation, the latter to evaluation manipulation, and the study directly examines the relation between the two dynamics. Connections synthesized across separate bodies of literature, rather than directly examined within a study, are presented as open research questions.

This mapping separates the function represented by each finding from the relation it may support, allowing individual studies to contribute to multiple parts of the taxonomy without treating the categories as mutually exclusive at the paper level.

\subsubsection{Literature collection and scope}

To assemble the evidence base, we conducted a structured literature search for work available through July 1, 2026. Query families aligned with the six taxonomy categories were applied across arXiv, OpenReview, ACL Anthology, PMLR, JMLR, publisher and proceedings pages, DOI records, and official conference websites. We complemented this search by tracing original studies from existing surveys and retrieving related work based on titles, abstracts, and topical similarity. We later conducted targeted updates for newly released work; as discussed in Section~\ref{sec:limitations}, these updates did not rerun every search query.

The search and retrieval process yielded approximately 500 candidate papers and institutional records. After deduplication and manual screening for relevance to AI-mediated scientific production, evaluation, and the interactions represented in our taxonomy, we retained 230 sources. Recent preprints were included when they provided directly relevant evidence, with publication status tracked separately. Work with only weak connections to the AI paper-review process was excluded unless it established a mechanism that clearly transferred to one of the six categories. For retained sources, claims used in the survey were checked against the corresponding original paper, report, policy, or official record whenever available.

Study settings and designs are described when they materially constrain interpretation. In particular, controlled demonstrations are not used to infer real-world prevalence, announced policies are not treated as evidence of effectiveness, and plausible downstream pathways are not presented as observed causal cycles.

Appendix Tables~\ref{tab:data-resources} and~\ref{tab:evidence-map}
complement this mapping by summarizing reusable scholarly records and
representative empirical evidence for the six dynamics and their response
relations.

\section{Scaling scholarly production}
\label{sec:scaling}
\label{sec:production}

Production scaling refers to how AI can enable research outputs to be produced in greater volume or in less time. By reducing the time and effort required for research production, revision, and resubmission, AI can enable researchers or automated systems to produce more outputs, complete them faster, or both. This is important because production capacity and evaluation capacity do not necessarily scale together.

\subsection{From industrialized production to AI-enabled scaling}

Large-scale production of scholarly artifacts is not entirely new. Before generative AI, its most visible problematic form was industrialized production through paper mills and related authorship-for-sale practices \citep{porter2024identifying,abalkina2023publication}. Such operations showed that research-like artifacts could be produced systematically when production was organized around volume rather than individual scholarly workflows. Generative AI changes the accessibility and scope of this capability. Tasks that previously required substantial individual time, including literature search, drafting, experimentation, and formatting, can increasingly be assisted or automated within ordinary research workflows.

The first visible change has been the widespread adoption of AI-assisted writing. Corpus studies find distributional changes in key sections of scientific papers that are consistent with substantial LLM-associated modification in computer-science preprints and biomedical publishing \citep{liang2025quantifying,kobak2025delving}. These studies confirm the diffusion of AI-assisted writing and indicate that AI-enabled writing tools have become widely used in scientific communication.

More capable research systems extend this shift beyond writing assistance. ResearchAgent supports iterative idea development over the literature; the AI Scientist and its successor integrate ideation, experimentation, manuscript production, and evaluation; and CycleResearcher uses automated review as a revision signal \citep{baek2025researchagent,lu2024ai,yamada2025ai,weng2025cycleresearcher}. Agent Laboratory, PaperOrchestra, and PaperOrchestrator further connect literature, code, experimental logs, and manuscript generation \citep{schmidgall2025agent,song2026paperorchestra,
yuan2026paperorchestrator}. Related systems extend these workflows to coding, bioinformatics, and multimodal medical research \citep{jansen2025codescientist,tang2026ai,ifargan2025autonomous,
kamran2026prompt,wu2026towards}. Other systems emphasize particular stages of the research process rather than full paper production. The AI co-scientist, for example, uses multiple agents to generate, debate, and refine hypotheses that are subsequently tested by human scientists in biomedical settings \citep{gottweis2025towards}. Together, these systems illustrate different levels of automation across the research workflow.

More importantly, the key change is not simply that individual systems contain more components, but that multiple stages of research can increasingly be coordinated within the same automated workflow. This makes it cheaper to explore multiple research directions, discard unsuccessful branches, and turn successful ones into complete artifacts. AgentRxiv, for example, allows agent laboratories to share intermediate reports so that research trajectories can be compared and reused rather than executed independently \citep{schmidgall2025agentrxiv}. EvoScientist and AutoResearchClaw similarly carry information across experiments or runs \citep{lyu2026evoscientist,liu2026autoresearchclaw}, while Jr.\ AI Scientist, PaperClaw, and Claw AI Lab introduce mechanisms such as provenance, experimental logs, rollback, and human intervention \citep{miyai2025jr,ye2026paperclaw,wu2026claw}.

Deployment-scale systems further show that this increase in throughput is no longer limited to small demonstrations. Fully Automated Research System (FARS) reports a public deployment that produced 166 complete papers across 67 AI/ML topics while preserving proposals, code, logs, results, and manuscripts \citep{tang2026fars}. This demonstrates that complete research artifacts can be generated at substantial scale, but producing a complete artifact is not the same as producing scientifically strong or successful research. In 282 structured reviews of 140 FARS outputs, the mean overall score was 3.17 on an ICLR-style scale, and only 11.4\% reached a mean score of six. Artifact-grounded audits also identified integrity problems, with insufficient experimental evidence emerging as a major weakness. A complementary result comes from two shadow evaluations in which frontier agents completed substantial engineering work and produced full manuscripts but did not solve the main open-ended research problems \citep{kirgis2026can}. Human-verifiable and laboratory-coupled systems provide alternative designs \citep{ifargan2025autonomous,ghareeb2026multi}, but the broader pattern remains the same: the capacity to produce research artifacts can scale faster than the capacity to establish that their scientific claims and conclusions are sound.

\subsection{Empirical evidence of faster and greater production}

Evidence from real scholarly corpora increasingly connects AI adoption with changes in research output. Across 2.1 million preprints from arXiv, bioRxiv, and SSRN, \citet{kusumegi2025scientific} estimate that inferred AI adoption is associated with output increases of 36.2\%, 52.9\%, and 59.8\%, respectively. Estimated gains vary substantially across author backgrounds and fields, and are particularly large for some researchers facing higher English-writing costs. The same study suggests that lower production barriers may also broaden access to different forms of literature and participation. However, adoption is inferred from text, and the design does not provide causal identification.

\citet{renault2026comment} question the size of this estimate. Because adoption is defined by an author's first detector-flagged paper, authors whose output is already rising are more likely to be classified as adopters. Their simulations and placebo tests recover similar post-adoption increases even when LLM use has no effect. The analysis does not show that LLMs have no productivity benefit; it shows that this design does not identify that benefit.

Other studies provide complementary evidence for changes in scholarly production. Using matched author panels in the social and behavioral sciences, \citet{filimonovic2025can} associate adoption with 15\% higher productivity in 2023 and 36\% in 2024, although adoption is again inferred from textual markers and residual selection remains possible. At the aggregate level, \citet{qi2025does} analyze more than 2.1 million records from four preprint repositories and find faster submission and revision cycles after the release of ChatGPT, particularly in computational fields. Their interrupted time-series design cannot fully separate the effect of LLM availability from other contemporaneous changes.

Taken together, these studies show that AI-assisted writing is widespread and that inferred adoption is associated with higher output or faster production cycles. They do not yet establish how much of this relationship is causal or how broadly it generalizes across fields and researcher groups.

\subsection{Scaling continues after submission}

Production does not end when a manuscript is submitted. Review, rebuttal, revision, and resubmission create additional production cycles, and AI can reduce the time and effort required at these stages as well. Response systems now decompose reviews into specific concerns, retrieve relevant manuscript evidence, plan responses, and learn from earlier response failures \citep{li2025self,han2026drpg}. Paper2Rebuttal and REspGen introduce structured plans, human checkpoints, author-specific information, and response controls, while XtraGPT applies context-aware revisions directly to the manuscript \citep{ma2026paper2rebuttal,ruan2026author,chen2026xtragpt}.

These systems can be beneficial when they help authors identify missing evidence, clarify existing results, or respond more efficiently to reviewers. Their significance for production scaling is that the reduction in time and effort extends beyond the initial submission. Reviewer feedback can be processed and converted into a revised manuscript or rebuttal more quickly, allowing the same submission to move through evaluation and response cycles at a faster pace.

The difficulty of post-submission revision depends on what the reviewer asks the author to change. Some revisions mainly reorganize or improve the presentation of existing work, while others require new experiments, code, or scientific reasoning. APRES focuses on the former setting, keeping the core scientific content fixed while optimizing presentation; experts often preferred its revised versions~\citep{zhao2026apres}. ReviseBench evaluates the harder setting, where agents must respond to reviews through writing, coding, and experimentation. Across 12 ICLR papers, model-generated revisions remained substantially behind human camera-ready versions and sometimes introduced incomplete experiments, formatting errors, or fabricated results~\citep{luo2026can}. Together, these studies suggest that AI is currently better at revising how existing work is presented than at carrying out the new scientific work required to address substantive reviewer concerns.

Production scaling after submission is therefore uneven: AI can substantially reduce the cost of responding to evaluation, while revisions that require new evidence, experiments, or scientific judgment remain harder to automate.

\subsection{From production scaling to evaluation pressure}

The downstream consequence appears when this greater volume and faster rate of iteration reach an evaluation process whose capacity remains limited. The clearest evidence comes from settings where submission activity and reviewing workload can be observed directly. At Organization Science, \citet{gartenberg2026more} analyze 6,957 initial submissions and 10,389 textual reviews from 2021--2026. Submission volume increased by 42\% after ChatGPT's release, with much of the increase concentrated in manuscripts receiving high aggregate AI-assistance scores, while the journal experienced additional editorial and review load. Because this is an observational study of one journal using a commercial detector, it does not establish a general causal effect. It nevertheless provides direct evidence that AI-associated production growth can translate into additional evaluation demand.

Evidence from ICLR points in the same direction. Using an extended submission series, \citet{knight2026flooded} estimate that annual submission growth accelerated from 25\% before ChatGPT to 59\% afterward. In a sample of 18,787 submissions, they also estimate that the relationship between acceptance and later citation weakened. The authors further propose two related mechanisms: increased volume spreads evaluator attention more thinly, while cheaper linguistic polish reduces the usefulness of presentation quality as a signal for triage.

Beyond increasing evaluation volume, cheaper production can also make redundant or overlapping work harder to identify. In a biomedical corpus, \citet{maupin2025dramatic} identify 411 effectively redundant papers among 3,465 studies of selected NHANES exposure-outcome pairs. In a separate demonstration, three syntactically different manuscripts could be generated in roughly two hours each while remaining below individual-source similarity warning thresholds. This illustrates a related problem: when producing alternative versions becomes cheap, evaluators may need to distinguish semantically redundant work even when surface-level similarity checks fail. Reports of reworded dual submissions provide a related conference-level example \citep{wagstaff2026aaai}.

\paragraph{Synthesis.}
In conclusion, the evidence shows that generative AI can make scholarly production both faster and more scalable, from initial research and manuscript preparation to later revision and rebuttal. Evaluation capacity, however, remains constrained by the time and expertise required to read, verify, and judge scientific work. Production scaling therefore shifts pressure downstream and creates growing demand to scale evaluation as well. The next section examines how AI is being used to support and automate scholarly evaluation.

\section{Automating scholarly evaluation}
\label{sec:evaluation}

As scholarly production scales, evaluation faces the same pressure to process more work with limited expert attention. AI offers one way to expand this capacity by supporting or automating parts of peer review. However, evaluation automation covers systems with very different levels of authority. An AI system may provide private feedback to authors, assist reviewers in preparing reports, contribute an official review, or support scoring, meta-review, and publication decisions. As AI moves closer to consequential decisions, the central question shifts from whether it can generate useful review text to whether its judgments are reliable enough for the authority it receives.

\subsection{From review tasks to evaluative roles}

The use of AI to scale evaluation builds on earlier computational work on peer review. PeerRead released 14.7K paper drafts, decisions, and 10.7K expert reviews, enabling tasks such as acceptance and aspect-score prediction \citep{kang2018dataset}. Later datasets supported review aggregation and meta-review generation \citep{shen2022mred,li2023summarizing,li2024sentiment}. With the emergence of LLMs, the range of automated tasks has expanded further to submission checking, reviewer-paper matching, review drafting, claim verification, rebuttal support, meta-review, and decision assistance \citep{kuznetsov2024can,zhuang2025large,wu2026can,du2024llms,hossain2025llms,yu2024automated}.

As this range expands, the more important distinction is not simply which task is automated, but how much evaluative authority the system receives. Author-facing feedback has no direct influence on publication decisions. Reviewer assistance may change an official report while leaving the reviewer responsible for the judgment. An additional AI review introduces a new evaluative voice into formal deliberation, while scoring, triage, and meta-review place the model closer to publication decisions. Capability at one level therefore does not automatically justify use at the next. Producing helpful feedback is different from identifying a fatal scientific flaw, and agreement with existing reviews is different from making a reliable publication decision.

\subsection{From assistance to official evaluation}

This progression in evaluative authority is already visible in current systems and deployments. The lowest-authority form of evaluation automation is feedback provided directly to authors. 
\citet{liang2024can} generated GPT-4 feedback for full manuscripts and compared it with reviews of 3,096 accepted papers across 15 Nature-family journals and 1,709 ICLR submissions. Model feedback showed overlap with human reviews, and 57.4\% of 308 researchers in a prospective sample rated it helpful or very helpful. The system nevertheless struggled with deeper methodological criticism, suggesting that automated pre-review can expand access to feedback without reproducing the full role of expert review.

Other author-facing systems combine planning, literature search, and critique to suggest revisions or identify concerns later raised by reviewers \citep{chamoun2024automated,wu2026drafting}. A NeurIPS 2024 deployment similarly used an LLM-based checklist assistant on 234 voluntary submissions from 184 papers. More than 70\% of post-use respondents found the tool useful, and a similar share intended to revise their paper or checklist \citep{goldberg2024usefulness}. Importantly, these systems remained outside the formal decision process.

Although author-facing feedback remains outside the formal decision process, it can still influence subsequent author behavior at scale. In a randomized experiment with 31,020 arXiv manuscripts, customized AI feedback increased the one-month revision rate by 12.55\% relative to the control group~\citep{wang2026human}. The effect was larger for authors with less access to established feedback networks. This shows that low-authority AI evaluation can affect how authors respond to their work even without directly influencing publication decisions.

Reviewer-facing assistance moves one step closer to official evaluation. At ICLR 2025, \citet{thakkar2026large} conducted a large-scale randomized intervention in which an AI system provided optional suggestions for improving review reports. Among 44,831 reports, 18,946 received feedback, and 5,031 were revised, incorporating 12,222 suggestions. Revised reports were judged clearer and more informative, and the intervention increased engagement during rebuttal. The AI system, however, did not write the official review, change scores, or enter the acceptance decision. The experiment therefore shows that AI can change formal review practice at scale while leaving the final evaluative authority with human reviewers.

AAAI-26 moved further along this authority ladder. Every one of the 22,977 main-track papers entering full review received a clearly labeled AI review generated in less than 24 hours \citep{biswas2026ai}. The system evaluated multiple aspects of the paper, used code and search tools, critiqued its own draft, and retained logs for oversight. Across 5,834 survey responses, participants often preferred the AI reports on technical accuracy, focus, and research suggestions.

The AI review still did not provide a score or acceptance recommendation and did not replace a human reviewer. It nevertheless became part of the official evaluation record and could be consulted by senior committee members. This deployment therefore marks an important transition: AI-generated evaluation has moved from private assistance into live conference workflows, even while venues continue to limit its formal decision authority.

\subsection{Toward grounded and auditable evaluation}

As AI receives greater evaluative responsibility, simply generating a plausible review becomes insufficient. Recent systems increasingly aim to make review judgments more structured, grounded, and auditable.

One direction is workflow decomposition. Instead of generating a review in a single step, systems divide evaluation into stages such as summarization, retrieval, critique, and decision. ReviewAgents follows a process of summarization, retrieval, critique, and decision, while Generative Agent Reviewers combines personas, claim graphs, simulated reports, and meta-review \citep{gao2025reviewagents,bougie2025generative}. PeerCheck similarly shows that review quality depends on how reasoning and retrieval are incorporated, with effects varying across models \citep{chen2026peercheck}. Reviewer guidelines also influence model judgments: official conference guidelines can move model scores closer to human averages, while reviewer-imitating or rigid additive guidelines may perform less consistently \citep{li2026evaluating}. The evaluation workflow is therefore part of the reviewer itself rather than a neutral wrapper around the model.

A second direction is grounding judgments in evidence. Novelty assessment requires comparison with external literature \citep{afzal2026beyond}, while scientific support requires checking whether claims are justified by evidence inside the manuscript \citep{ballakuraya2026methods}. Review grounding further asks whether a criticism is supported, contradicted, or left unresolved by the paper \citep{ghorbanpour2026peerispect}. ReviewGrounder and ScholarPeer combine paper-specific criteria with retrieval and specialist agents to support these functions \citep{li2026reviewgrounder,goyal2026scholarpeer}. The broader goal is to make criticism depend less on surface plausibility and more on traceable scientific evidence.

A third direction is auditability. EGTR-Review retains retrieved evidence, sources, and uncertainty while generating reviews \citep{qiu2026egtr}. ProReviewer represents review as a sequential investigation whose log records claims, questions, and observations that guide later inspection \citep{fang2026passive}. DeepReviewer 2.0 similarly produces a claim-evidence-risk record with anchored criticism and repair actions \citep{weng2026deepreviewer}. These systems move automated review away from reviewer imitation and toward evaluations whose supporting evidence and intermediate judgments can be inspected.

\subsection{Automation already affects selection}

Formal deployments capture only part of current evaluation automation because reviewers can also use general-purpose LLMs independently. A distributional analysis of ICLR, NeurIPS, CoRL, and EMNLP reviews estimates that 6.5--16.9\% of post-ChatGPT review text was substantially AI-modified \citep{liang2024monitoring}. This suggests that AI-assisted evaluation entered review practice even before many venues established official systems.

More importantly, AI-associated reviewing is already linked to selection outcomes. \citet{russo2025ai} estimate that at least 15.8\% of ICLR 2024 reviews were AI-assisted. Within-paper comparisons find these reports slightly more lenient, while matched analyses associate receiving an AI-assisted review with a 3.1 percentage-point increase in acceptance and a 4.9-point increase for borderline papers. Although the study relies on inferred AI use rather than direct observation, it shows that AI-associated reviewing can coincide with consequential differences in evaluation outcomes.

A larger analysis of more than 125,000 paper-review pairs further suggests that automation can change calibration and information aggregation \citep{sharma2026llms}. Reviews classified as LLM-associated show greater leniency toward weaker papers, while fully generated reviews in synthetic experiments exhibit stronger rating compression. LLM-associated meta-reviews are also more likely to recommend acceptance at equivalent reviewer scores. These findings indicate that evaluation automation can affect not only how reviews are written, but also how judgments are distributed and combined.

\subsection{The reliability bottleneck}

Despite rapid progress in system design, reliable scientific judgment remains the central limitation. Specialized models, multi-agent workflows, retrieval, question decomposition, and reinforcement learning continue to improve review generation \citep{idahl2025openreviewer,d2024marg,zhu2025deepreview,chang2025treereview,zeng2025reviewrl}. However, conventional evaluation of AI reviewers often measures fluency, overlap with human reviews, or general helpfulness rather than whether the system correctly identifies and prioritizes scientifically important problems \citep{zhou2024llm}.

PRISM evaluates automated reviewers across analytical depth, novelty verification, flaw identification and prioritization, and constructiveness \citep{loc2026prism}. Individual systems perform strongly on selected dimensions, but no single system reproduces the same balance across all of them. ReviewEval similarly evaluates multiple dimensions of generated reports \citep{garg2025revieweval}, while CLAIMCHECK asks whether a criticism is actually grounded in the scientific claim it challenges \citep{ou2025claimcheck}. Together, these benchmarks shift attention from whether a review sounds plausible to whether its judgments are scientifically supported.

Results on criticism-level tasks are more mixed. In a study of 82 Nature-family papers with 45 expert scientists, GPT-5.2 outperformed the top-rated human review on a composite measure of correct, important, and well-supported criticisms (60.0\% versus 48.2\%)~\citep{kim2026limits}. However, its individual criticisms were less often correct, and AI reviews overlapped much more with one another than human reviews. These results show that strong performance on criticism generation can coexist with lower per-criticism reliability and less diversity across reviews.

Strong criticism generation, however, does not necessarily imply sensitivity to scientific soundness. Counterfactual evaluation reveals a sharper limitation. Across 133 accepted AI and NLP papers, \citet{dycke2026automatic} construct 931 variants, including 391 edits that break critical scientific support relations and 540 soundness-neutral controls. Across the tested automated reviewers, soundness-critical changes did not produce statistically significant differences in review aspects, sentiment, or scores compared with surface-level control edits. The models were also sensitive to irrelevant wording changes.

Direct error-detection benchmarks point to the same verification gap. On SPOT, which contains 91 author-validated errors from 83 published papers, no tested model exceeded 21.1\% recall or 6.1\% precision~\citep{son2025ai}. The models also rarely found the same error across repeated runs. In MLReplicate, automated review accepted 10 of 37 valid agent-generated papers, and 59\% of the accepted papers contained fabricated or unsupported claims~\citep{gaddipati2026mlreplicate}. The benchmark covers only eight machine-learning tasks, but it shows that weak generation and weak evaluation can compound each other.

These findings narrow the evaluation bottleneck. Generating criticism is becoming cheap, and AI can perform well on some bounded tasks. The harder problem is to verify claims, identify the flaws that matter most, preserve different perspectives, and make accountable decisions.

\paragraph{Synthesis}
Evaluation automation can expand review capacity, and its role has already progressed from private feedback to assistance within formal review and official AI-generated reports. At the same time, greater authority places greater demands on the reliability of scientific judgment, which remains difficult to automate. Machine-mediated evaluation also makes parts of the review process more repeatable and easier to query. Once authors can observe and respond to these regularities, the evaluator itself becomes a target for optimization. Section~\ref{sec:manipulation} examines this next dynamic.

\section{Evaluation manipulation}
\label{sec:manipulation}

As AI-mediated evaluation becomes more repeatable and queryable, participants can begin to exploit its regularities to influence how their work is evaluated. This creates the third dynamic in our taxonomy: evaluation manipulation. Such manipulation can directly target an AI reviewer through hidden instructions, but it can also exploit presentation, framing, or other evaluator-visible signals that systematically affect review outcomes. Repeated access to evaluator feedback can further turn these signals into targets for iterative optimization.

\subsection{Defining evaluation manipulation}

Not every revision that improves an AI review is manipulation. Clearer writing, corrected analysis, additional evidence, or a better explanation of limitations can improve both the paper and its evaluation. The relevant distinction is whether a change primarily improves the scientific work or instead targets how the evaluator responds to it. 
Importantly, evaluation manipulation differs from evasion in both its target and its position in the response sequence. Manipulation targets the evaluator in order to shift its judgment without a corresponding improvement in the underlying scientific evidence. Evasion instead occurs after a defense or restriction is introduced and adapts behavior to avoid or circumvent that control. The same technique may therefore count as manipulation when it directly targets an evaluator, and as evasion when it is adopted in response to a safeguard.

This distinction is clearest for direct instructions to the reviewer, but becomes less obvious for presentation changes. Repositioning related work may genuinely clarify novelty or simply make the same contribution appear stronger. A revised discussion may explain a limitation more clearly or reduce the weight that an evaluator assigns to an unchanged weakness. Controlled comparisons that hold methods, data, figures, equations, and numerical results fixed while changing presentation provide one way to study this boundary \citep{yang2026no,li2026gaming}. The key question is therefore not simply whether a score changes, but what was changed and why the evaluator responded differently.

Controlled revision studies further illustrate this distinction. LLM-REVal finds that iterative model feedback can produce genuine corrections while also exposing preferences that authors can exploit, and Review Arcade shows that repeated revision against an LLM reviewer does not consistently improve evaluation across cases \citep{li2025llm,hatzel2026review}. A higher score alone does not show that a revision improved the science or manipulated the reviewer. The relevant distinction is whether a revision improves the work or its assessability, or instead changes the evaluator's judgment without a corresponding change in what the scientific evidence supports.

\subsection{From explicit to subtle manipulation}

The clearest form of evaluation manipulation directly targets the AI reviewer. A July 2025 audit identified 18 arXiv preprints containing concealed instructions, often presented as white or microscopic text, asking an AI reviewer to provide favorable assessments or particular review content \citep{lin2026hidden}. These observations show that attempts to manipulate AI-mediated review have already appeared in public manuscripts. Related work shows that such instructions can also be concealed through font encodings, multilingual text, or external resources \citep{collu2026misleading,choi2026invisible,
theocharopoulos2025multilingual,xiong2025invisible}.

Controlled studies show that these attacks can affect AI-review outcomes. Hidden instructions have increased ratings and reduced reported weaknesses across multiple review systems \citep{ye2024we,zhu2025your}. Larger evaluations also reveal substantial variation across models: some tested reviewers remain highly vulnerable, while others are considerably more resistant \citep{li2026llm}. That is to say, a reviewer that is robust in one setting may still be vulnerable under a different model, prompt, or evaluation pipeline.

Manipulation also extends beyond a literal hidden sentence. Placement, layout, and domain-specific instructions can change attack effectiveness \citep{torrielli2026exploiting,sahoo2025reject}, while character-, word-, and sentence-level perturbations can influence reviewer behavior without remaining recognizable as explicit instructions \citep{lin2025breaking}. Scientific papers are also multimodal documents. PaperGuard shows that attacks can operate through figures as well as text, with figure-only perturbations changing aggregate reviewer scores in its benchmark \citep{zhao2026does}. The relevant attack surface therefore includes the broader document presented to the evaluator, not only visible prose.

More importantly, evaluation manipulation does not require an explicit instruction. Meaning-preserving rewrites can also change AI-review outcomes. Abstract-level paraphrasing has produced score gains across different papers and reviewers while preserving the intended meaning of the text \citep{li2026gaming,kaneko2026paraphrasing}. Full-paper experiments make this effect more direct. \citet{yang2026no} hold scientific evidence fixed while iteratively changing presentation based on AI-reviewer feedback. Across three reviewer models, the attack achieved a 75.1\% success rate and increased scores by an average of 1.21 points on a ten-point scale. Changes to related-work positioning and analytical discussion were particularly effective, and criticisms of unchanged limitations sometimes weakened or disappeared.

\citet{baumann2026stop} report a similar effect: rewriting papers without changing their scientific content increased AI-review scores by an average of 0.45 points. They also find that AI reviews are more similar to one another than human reviews, suggesting that susceptibility to presentation changes may coexist with reduced diversity in evaluation.

These results show a progression from explicit instructions to ordinary-looking changes in presentation. An evaluator may still recognize the same scientific weakness while changing how much that weakness affects its overall judgment. Related work on human review and LLM judges similarly identifies sensitivity to citation, position, rubrics, style, confidence, and framing \citep{stelmakh2023cite,wang2024large,ding2026rubrics,yang2026turning,wang2026ai}. Such sensitivities are not necessarily manipulation by themselves, but they create predictable signals that can be exploited.

A more severe adjacent failure occurs when convincing presentation is combined with unreliable or fabricated scientific content. BadScientist shows that agent-generated manuscripts without real experiments can nevertheless receive favorable assessments from LLM reviewers under controlled conditions \citep{jiang2026badscientist}. This differs from presentation-based manipulation, but it reinforces the broader concern that an evaluator can assign favorable judgments when plausible presentation is not matched by sufficient scientific verification.

\subsection{From one-shot to adaptive manipulation}

The more consequential shift occurs when manipulation becomes iterative. A one-time hidden instruction demonstrates that an evaluator can be manipulated. A repeatable evaluator, however, also reveals whether a particular change worked. The manuscript can then be revised, evaluated again, and further adjusted based on the new response.

Controlled prompt-injection studies already demonstrate this process. Iterative attacks use reviewer scores or surrogate-model feedback to refine hidden instructions, producing stronger attacks that can transfer across evaluators \citep{zhou2025give}. The same principle applies to presentation. Repeated reviewer feedback can guide which sections to rewrite, which framing to emphasize, and which changes are most likely to alter the eventual assessment \citep{li2026gaming,yang2026no}.

This process does not require exact knowledge of the deployed evaluator. Aspect-level perturbation studies reveal systematic sensitivities in review models, while work on generic LLM judges shows that scalar evaluator feedback can be used to optimize transferable prompts without corresponding improvements in human judgments \citep{li2025llms,raina2024llm,shi2024optimization,alazraki2026reverse}. Keeping the deployed model or prompt secret can make optimization more difficult, but shared preferences and cross-model transfer can still provide useful signals.

The central mechanism is therefore straightforward: repeated evaluation turns reviewer regularities into an optimization signal. Once this happens, manipulation is no longer limited to a fixed hidden prompt or a single rewrite. Authors or automated systems can search over instructions, wording, framing, layout, and other evaluator-visible features and retain changes that produce more favorable evaluations. Evaluation manipulation can therefore become adaptive rather than one-shot.

\paragraph{Synthesis.}
The evidence shows that AI-mediated evaluation can be manipulated through both explicit attacks and more subtle changes to evaluator-facing signals. Repeated access to evaluator feedback can further turn these sensitivities into an adaptive optimization process. Evidence that such manipulation changes official venue decisions remains limited, but the underlying mechanisms are already observed and experimentally reproducible. These risks create pressure for venues to protect the reliability of evaluation, motivating the technical and institutional defenses examined next.

\section{Defense mechanisms and policy responses}
\label{sec:defense}

As evaluation manipulation becomes observable and reproducible, defenses have expanded from detecting suspicious AI-generated content or hidden instructions to more targeted verification and more robust evaluation pipelines. These approaches aim not only to identify manipulation after it occurs, but also to reduce the vulnerability of AI-mediated evaluation itself. At the same time, conferences, journals, and publishers are increasingly incorporating such safeguards into review policies and procedures that define permitted AI use, accountability, verification, and enforcement.

\subsection{From AI detection to targeted verification}

A natural first response to AI-mediated risks is to detect whether AI was used. General-purpose AI-text detectors and review-specific classifiers attempt to distinguish human-written from AI-generated or AI-assisted text. In peer review, paper-conditioned similarity and review-specific methods can improve detection under low false-positive requirements \citep{yu2024your,duarte2026sem}. However, these systems infer an unobserved writing process from the final text, and their performance changes across models, editing styles, and operating thresholds.

Real venue use illustrates this limitation. In a NeurIPS 2026 position-paper audit, a default document-level detector marked 28.2\% of 969 submissions with the maximum AI score, while a 100-word-window analysis reduced the high-confidence share to 12.7\% \citep{neurips2026aigeneratedpapers}. The large difference shows that detector outputs depend strongly on how the artifact is segmented and scored. Human judgment is also not a reliable substitute: experienced reviewers can struggle to distinguish human from AI-augmented writing \citep{hadan2024great}.

Detection becomes particularly difficult when policies allow some forms of AI assistance but prohibit others. On a dataset of 50,156 reviews spanning human and simulated human-AI collaboration, commercial detectors identified most fully generated reviews but also classified some compliant AI-polished reviews as fully AI-generated \citep{saha2026policies}. This makes post-hoc attribution useful for screening, but much weaker as direct evidence of how a particular review was produced.

These limitations motivate a shift from passive detection toward targeted verification. Instead of asking whether a text merely looks AI-generated, a venue can introduce a known signal and test whether a prohibited process actually occurred. \citet{rao2025detecting}, for example, randomly assign covert manuscript markers that cause an LLM generating a review to emit a rare phrase or citation. Random-citation markers appeared in 98.6\% of tested LLM reviews on average and often survived one-shot paraphrasing. Unlike ordinary stylometry, this approach creates a controlled probe of a specific behavior rather than inferring provenance from writing style.

This idea has already moved into practice. An ICML 2026 report describes the use of randomized manuscript watermarks followed by manual verification and sanctions for reviewers who had agreed to a no-LLM policy \citep{icml2026llmreviewviolations,saha2026policies}. The mechanism does not detect every form of AI assistance, but it illustrates a broader change in defensive design: when attribution from text is unreliable, venues can verify narrower behaviors using controlled and auditable signals.  

Related approaches extend targeted verification beyond randomized watermarks. IntraGuard embeds refusal instructions or markers in manuscripts to expose end-to-end LLM review outsourcing \citep{ma2026shattering}, while other work studies watermark, refusal, and monitored-redirection mechanisms under different attack settings \citep{nemecek2025feasibility,torrielli2026exploiting}. These methods can be effective against the behaviors they target, but rephrasing, modality changes, or jailbreaks can weaken them. Together, they reinforce the shift from inferring AI use from writing style toward testing specific behaviors through controlled signals.

\subsection{Protecting AI-mediated evaluation}

Detection alone does not protect an AI reviewer from manipulation. A second line of defense therefore focuses directly on the path from manuscript to evaluation.

The first step is to control the input presented to the evaluator. Hidden or microscopic text can be identified by comparing rendered and extracted content, suspicious metadata and external resources can be isolated, and the manuscript can be treated as untrusted data rather than as an extension of the reviewing prompt. These measures can block common prompt-injection channels, although they cannot address manipulation that operates through legitimate-looking presentation alone.

More targeted systems explicitly test the evaluator against adversarial inputs. A detection prompt in \citet{zhou2025give} identifies many non-adaptive hidden instructions and moves reviewer scores back toward their baseline, but its effectiveness drops when the attacker is allowed to adapt to the defense. SafeReview addresses this problem by jointly exposing the reviewer to progressively stronger attacks during training \citep{xin2026safereview}. Under the evaluated attacker configurations, it preserves paper ranking better than static adversarial training and avoids some of the excessive blocking produced by generic prompt detectors. The broader lesson is that defenses against adaptive manipulation must themselves be evaluated under adaptive attacks.

Long and multimodal papers create additional difficulties. PaperGuard searches manuscript passages and figures for instruction-like content and then verifies the retrieved candidates \citep{zhao2026does}. Artifact-specific methods similarly target fabricated scientific tables or figures \citep{huang2026tab,hu2026scifigdetect}. These tools broaden the protected surface beyond plain text, but they also illustrate the limit of attack-specific detection: presentation-level manipulation may contain no discrete malicious string or artifact to identify.

Defenses may also need to examine the research process, not just the final paper. Verification-first proposals argue that AI should produce evidence and checks that can be audited, rather than imitate review scores or final decisions~\citep{you2026preventing}. A controlled study supports this direction: adding code, logs, and execution traces raised audit accuracy from 55\% to 82\% and exposed errors that were hard to see in the paper alone~\citep{luo2025more}. These results extend defensive verification beyond the submitted manuscript to the evidence and intermediate artifacts that support it.

\subsection{Institutional and policy responses}

Technical safeguards operate within venue rules that determine which uses of AI are allowed and how violations are handled. Current policies vary widely. Some venues prohibit reviewers from uploading confidential manuscripts to external models, some permit limited language assistance, others require disclosure, and some provide bounded venue-controlled tools while keeping human reviewers responsible for the final report \citep{iclr2026llmpolicy,arrreviewerguidelines,aclpublicationethics,
emnlp2026aireviewing}. Similar variation appears across journals: among the top 100 medical journals, 78 provided AI guidance in 2024, with policies ranging from prohibition to limited permitted use \citep{li2024use}.

Policy studies show large differences in whether guidance exists, where it appears, and what authors must disclose. In 2023, 24\% of the 100 largest publishers and 87\% of 100 highly ranked journals had generative-AI guidance~\citep{ganjavi2024publishers}. Another study found public AI policies at 50 of 59 disaster-research journals; all allowed some AI use but imposed different conditions~\citep{armitage2026artificial}. Guides and expert-consensus work generally call for clear disclosure, human verification, and author responsibility~\citep{cleland2026and,fettiplace2026recommendations}. These studies describe policy design, not policy effectiveness.

Conference policies make these differences concrete. AISTATS 2026 permits author-side AI use under human responsibility but prohibits reviewer-side LLM use and treats favorable hidden instructions as misconduct. COLM requires disclosure of substantive AI assistance by authors and reviewers, while KDD combines review disclosure with restrictions on sending manuscript text to external services \citep{aistats2026cfp,colm2026cfp,kdd2025researchcfp}. Other venues combine automated screening with human follow-up: EMNLP routes integrity flags to senior verification, while ICLR uses automated triage together with multi-person checking and appeals \citep{eccv2026reviewerfaqs,emnlp2026paperintegrity,iclr2026llmresponse,
iclr2026reviewretrospective}.

Venues are also beginning to constrain the role AI plays rather than treating automation as all-or-nothing. Author-facing preflight systems have checked formal requirements and presentation before review, while reviewer-facing tools such as the ICLR feedback system and ARR's self-hosted assistant operate within deliberately limited roles \citep{jayaram2026towards,ohta2026ai,chen2026happens,arr2026revas}. These deployments keep particular inputs and outputs under venue control and preserve an identifiable human decision point.

Whether these policies materially change author behavior remains less clear. Comparative evidence finds similar growth in detected AI-assisted writing across journals with and without AI policies, but the causal effect of policy adoption is difficult to establish without adoption dates and pre-policy trends \citep{he2026academic,wang2026policy,he2026reply}. Current evidence therefore documents the spread of AI policies more clearly than their effectiveness.

Institutional responses also address the pressure created by scaling production. Submission limits, reciprocal-review requirements, desk screening, and workload-aware assignment can reduce evaluation demand or increase review supply without requiring a venue to determine whether a submission was AI-generated \citep{cvpr2025changes,aistats2026cfp,colm2026cfp,kdd2025researchcfp}. These mechanisms can relieve capacity pressure, but they can also introduce trade-offs in access, workload, and fairness \citep{cao2025dissecting}.

\paragraph{Synthesis.}
Defense mechanisms have evolved from passive AI detection toward targeted verification, more robust evaluation pipelines, and formal policy responses. Technical safeguards can identify suspicious behavior or reduce evaluator vulnerability, but they increasingly operate within venue-controlled processes in which human reviewers, chairs, or editors retain responsibility for consequential decisions. Conferences and publishers therefore combine technical safeguards with bounded AI roles, disclosure requirements, verification, and enforcement procedures. These interventions also change the costs and incentives faced by participants, creating opportunities for evasion and new side effects examined in the next section.

\section{Evasion and side effects}
\label{sec:evasion}

Defenses change the costs and constraints faced by participants, but they do not necessarily remove the underlying incentive to use or optimize AI. Once a particular signal, tool, or behavior becomes subject to detection or restriction, participants can adapt by changing how that behavior is carried out or how it appears to the evaluator. At the same time, defensive measures can create unintended consequences even when no strategic evasion occurs. We therefore distinguish two related outcomes: \emph{evasion}, in which actors adapt in response to a control, and \emph{side effects}, in which the control itself shifts errors, labor, or risk elsewhere in the scholarly process.

\subsection{Evasion after defensive intervention}

A defense can itself become a target once its output has consequences. AI-text detectors, for example, rely on statistical patterns that distinguish generated from human-written text, but these patterns can change across models, decoding strategies, domains, and editing processes. Large-scale benchmarks show substantial performance degradation under distribution shifts and adversarial modification, while studies of scientific writing similarly find that relatively small changes in wording or form can alter detector outputs \citep{dugan2024raid,odri2023detecting,taloni2024modern,liu2024great,matsubara2025large,shahriar2025erosion}. Once detection affects screening or enforcement, users therefore have an incentive to reduce or remove the signals on which it relies.

Controlled studies demonstrate many ways to do so. Post-generation methods use paraphrasing, rewriting, watermark modification, and transferable substitutions to change detectable traces while preserving much of the original content \citep{krishna2023paraphrasing,sadasivan2025can,wang2024raft,fang2025your,dathathri2024scalable}. Other approaches adapt generation itself through evasive prompts, detector-guided generation, style optimization, humanization, or semantics-constrained optimization \citep{kumarage2023reliable,zhou2024humanizing,wang2025humanizing,meng2025gradescape,pedrotti2025stress,gu2026mash,wang2026detector}. Peer-review-specific detectors provide more specialized signals, but mixed human-AI writing, paraphrasing, and new generators continue to make attribution difficult \citep{kumar2024quis,kumar2025mixrevdetect,chen2026coconuts,saha2026policies}. Similar rewriting can also weaken plagiarism and provenance signals while preserving substantial semantic reuse \citep{gupta2025all,hassanipour2024ability,murdock2025can,dilber2026impact}. Across these settings, the common pattern is that the underlying content can remain similar while the signal targeted by the defense changes.

Stronger defenses may in turn create new opportunities for adaptation. Randomized manuscript probes, for example, provide more direct evidence of particular forms of LLM review delegation than ordinary text detection \citep{rao2025detecting}. Once such mechanisms are known, sanitization, rewriting, alternative document-processing routes, or different models become plausible responses. Related studies on presentation and multimodal manipulation further show that evaluator-facing behavior can move beyond the particular textual signals targeted by a defense \citep{yang2026no,zhao2026does}. These results point to plausible directions in which future evasion may develop.

This distinction also highlights an important gap in the current literature. Controlled experiments provide substantial evidence that many defensive signals can be circumvented, but direct evidence of post-policy adaptation in real scholarly settings remains limited. Existing studies rarely follow the full sequence in which a venue introduces a specific defense, participants change their behavior in response, and the venue subsequently adjusts its safeguards. The mechanisms of evasion are therefore increasingly well understood, while longitudinal evidence of defense-driven counter-adaptation remains an open empirical question.

\subsection{Side effects of defensive measures}

Defensive measures can also create costs even when participants are not actively trying to evade them. AI-content detection provides the clearest example. False positives and uneven error rates can cause legitimate human or AI-assisted writing to be flagged, with particular concerns for non-native-English writers and other populations whose writing differs from a detector's training distribution \citep{liang2023gpt,pratama2025accuracy,boonsuk2026editor,ali2026stigmatization}. When detector outputs are incorporated into editorial or integrity processes, these errors can become more than classification mistakes.

Automated evaluation can introduce a separate source of unequal treatment. Controlled studies show that LLM reviewers may respond to evaluator-irrelevant author information such as institutional affiliation, seniority, or publication history \citep{vasu2026justice}. Detection bias and evaluation bias arise at different stages of the process, but both illustrate the same broader concern: automation can redistribute errors rather than simply reduce human workload or inconsistency.

The same pattern applies to labor and risk. AI assistance may reduce some drafting effort while creating new work in checking generated content, verifying evidence, resolving disagreements, investigating suspicious cases, or reviewing appeals. Studies of AI-assisted reviewing report reductions in perceived workload alongside continued verification and oversight requirements \citep{chen2025envisioning,ebadi2025exploring}. The use of external AI services can also create confidentiality concerns when unpublished manuscripts or review materials leave venue-controlled infrastructure. Automation can therefore shift work and risk to reviewers, chairs, editors, integrity staff, or authors rather than eliminating them.

Taken together, these findings show that the effects of a defense extend beyond whether it successfully detects or prevents the targeted behavior. Even a technically effective intervention can introduce false positives, additional verification work, confidentiality risks, or unequal burdens across participants. Defensive measures should therefore be evaluated not only by how well they address the original threat, but also by the costs and risks they create elsewhere in the review process.

\paragraph{Synthesis.}
Current evidence supports both sides of this dynamic. Controlled studies show that detection and other technical defenses can be circumvented through rewriting, generation-time adaptation, tool switching, and changes in the signals presented to evaluators. At the same time, deployed and observational evidence shows that defensive measures can introduce false positives, unequal errors, additional human work, and confidentiality risks. What remains under-studied is the full longitudinal sequence from a deployed defense to observed participant evasion and a subsequent institutional response. Defenses therefore do not simply remove risk: they can change both participant behavior and where the remaining costs are borne. These effects can also persist beyond a single review cycle through the papers, reviews, and decisions that enter the scholarly record, which we examine next.

\section{Long-horizon ecosystem feedback}
\label{sec:feedback}
\label{sec:record-feedback}

The preceding dynamics largely unfold within a single research and review cycle. Long-horizon ecosystem feedback begins when the artifacts produced in that cycle persist and are reused by later research or evaluation systems. Papers, reviews, rebuttals, decisions, corrections, and citations can enter searchable corpora, retrieval systems, training pipelines, scientific agents, and AI evaluators. Through this reuse, information and judgments produced in one round can influence the inputs, objectives, or behavior of later systems. The key question is therefore not simply whether AI-associated material enters the scholarly record, but whether that record becomes part of future production and evaluation.

\subsection{Scholarly artifacts enter the persistent record}

The first requirement for long-horizon feedback is persistence. Corpus studies and targeted audits show that AI-associated material is already entering searchable and published scholarship. Large-scale analyses identify substantial post-2022 changes in scientific writing associated with LLM use, while more targeted studies document visible generated fragments in indexed publications \citep{schmidt2024using,siler2026diffusion,strzelecki2025my,haider2024gpt,yao2026ai,spinellis2025false}. These studies differ in how they identify AI involvement, but together they establish a basic point: the contemporary scholarly record increasingly contains material shaped by generative AI.

Errors can persist as well. Studies of generated scientific writing report fabricated or inaccurate references, and large corpus audits identify nonexistent citations in published or publicly available research records \citep{walters2023fabrication,bhattacharyya2023high,topaz2026fabricated,zhao2026llm}. Once such references are indexed or repeated by later papers, they can acquire the appearance of a citation history and become harder to correct across the broader record \citep{camp2025citation}. Long-horizon feedback therefore begins with a simple but consequential fact: both useful AI-assisted content and identifiable errors can survive beyond the local research or review cycle.

\subsection{Scholarly records become inputs to future systems}

Persistence matters because scholarly artifacts are increasingly reused by AI systems. Published papers can become part of pretraining corpora or inference-time retrieval systems. Dolma, for example, contains tens of millions of scholarly papers used in the documented training mixture for OLMo, while OpenScholar retrieves from a large collection of open-access papers to support literature synthesis \citep{soldaini2024dolma,groeneveld2024olmo,asai2026synthesizing}. Other open corpora similarly incorporate sources such as arXiv and PubMed, making scholarly material an explicit component of later model development \citep{kandpal2026common}. Studies of memorization further show that identifiable information acquired during continued pretraining can persist through subsequent model adaptation \citep{li2026memorization}. These systems are generally designed to improve scientific reasoning and access to knowledge, but they also make the dependency on prior scholarly records explicit.

Reviews and publication outcomes provide an even more direct feedback channel because they encode judgments rather than scientific content alone. Peer-review datasets have long made papers, reviews, and decisions available for computational modeling \citep{kang2018dataset,yang2026paper}. More recent systems train directly on large collections of conference reviews or use review-rebuttal interactions and author responses as supervision for future feedback generation \citep{idahl2025openreviewer,wu2026rbtact,mun2026goodpoint}. In this setting, historical reviews can teach future evaluators what to criticize, how to score, and which forms of feedback are associated with author response.

Some systems already connect these records back to future production or evaluation. SPARK trains an evaluator on large-scale OpenReview data and uses it to filter newly generated research ideas \citep{sanyal2025spark}. Historical review patterns therefore influence which future ideas are retained before they become papers. ReviewGuard uses papers, reviews, decisions, and later citation outcomes to train and optimize a future evaluator \citep{rasool2026reviewguard}. Although citation is an imperfect measure of scientific value, the system demonstrates the relevant interface: outcomes from one round of scholarly evaluation can become objectives for a later evaluator. A related system extends this pathway back into research production. Scientific Judge learns from citation-based preferences, and Scientific Thinker then uses that evaluator as a reward model for generating new research ideas~\citep{tong2026ai}. This provides a more direct record-to-evaluation-to-production pathway, although the proposed ideas are evaluated primarily by LLMs.

These examples establish that scholarly records are not merely archives. Papers can become future model context, while reviews, responses, decisions, and citations can become supervision or optimization signals. The reuse pathway required for long-horizon feedback is therefore already operating in several production and evaluation systems.

\subsection{From reuse to ecosystem feedback}

Reuse creates the possibility that particular errors, manipulations, or broader statistical patterns propagate into later systems. Controlled studies show that even a small number of strategically inserted artifacts can sometimes have disproportionate downstream influence. In a biomedical knowledge-graph experiment, a single malicious abstract substantially altered downstream drug-disease rankings \citep{yang2024poisoning}. Related work on retrieval and model training shows that injected or synthetic information can redirect RAG outputs, alter medical model claims, or dominate retrieval results \citep{zou2025poisonedrag,alber2025medical,yu2026retrieval}. These studies do not demonstrate that manipulated scholarly papers have already produced the same effects in deployed scientific systems, but they establish mechanisms through which persistent artifacts could influence later retrieval and reasoning.

At a broader level, repeated reliance on AI may also shape the distribution of future research. \citet{messeri2024artificial} highlight the risk that heavy reliance on AI could narrow research questions, methods, and viewpoints while creating an illusion of understanding. Controlled studies provide related mechanisms: repeated training on generated content can amplify existing patterns and reduce diversity~\citep{shumailov2024ai,alemohammad2024self,seddik2024bad,guo2024curious,doshi2024generative}, while other work identifies correlated errors and increasing similarity across models~\citep{kim2025correlated,jiang2026artificial}. Evidence closer to scholarly production is observational: AI use has been associated with changes in citation patterns and collective topical breadth~\citep{mccreery2026llm,hao2026artificial}. Together, these studies identify plausible mechanisms and early signals of distribution-level feedback, but they do not establish that scholarly publishing has entered a closed recursive degradation cycle.

The central empirical gap is therefore provenance across generations of systems. Current evidence shows that AI-associated material enters the scholarly record, that papers and reviews are reused by later systems, and that such reuse can influence retrieval, training, idea selection, and evaluation objectives. What has not yet been traced is a complete artifact-level path in which a specific AI-mediated paper or review enters a known corpus, materially changes a named successor system, and then alters a later paper, review, or publication decision. Establishing such lineage remains an important open empirical problem.

\paragraph{Synthesis.}
Long-horizon ecosystem feedback is therefore partially established rather than fully closed. Record ingress is observed, and the reuse of scholarly papers, reviews, responses, and outcomes in later AI systems is already implemented. Controlled and adjacent studies further show mechanisms through which reused artifacts or repeated patterns can influence later systems. What remains missing is end-to-end evidence linking an identified artifact through a specific successor system to a later scholarly outcome. The final dynamic is thus best understood as a documented exposure and reuse pathway whose complete causal feedback loop remains to be demonstrated.

\section{Cross-cutting findings and research agenda}
\label{sec:findings}

All the previous dynamics reveal several properties of AI-mediated scholarly publishing that are difficult to see when production, evaluation, manipulation, and governance are studied separately. Three findings recur across the evidence: trustworthy evaluation remains harder to scale than research production, static evaluations become less informative once participants can adapt to AI-mediated review, and safeguards can shift rather than eliminate errors, labor, and risk. These findings also identify where the current evidence remains incomplete and where future research is most needed.

\subsection{Trustworthy evaluation remains the bottleneck}

Generative and agentic systems can increasingly reduce the time and effort required to produce, revise, and evaluate research artifacts. The corresponding work of verifying scientific claims, identifying consequential flaws, assessing novelty, and making accountable decisions remains much harder to scale. Evidence from journals and conferences already connects growing production volume with pressure on editorial and review capacity, while automated review systems can generate feedback at substantially lower marginal cost \citep{gartenberg2026more,knight2026flooded,biswas2026ai}.

How humans and AI are combined also matters. In a randomized study of 103 teams assessing research reproducibility, human-only and AI-assisted teams achieved similar performance (94\% and 91\%), whereas AI-led teams achieved about 37\%~\citep{brodeur2026ai}. Human-only teams also found more coding errors. The result further suggests that increasing AI involvement does not automatically reduce the need for human control in verification-intensive tasks.

The resulting bottleneck is therefore not simply review generation, but trustworthy scientific evaluation. Faster review generation is useful only when the resulting judgments remain grounded and when the human effort required for verification, disagreement resolution, and oversight does not grow at the same rate. Future deployments should therefore measure evaluation capacity in terms of both throughput and the human work required to maintain reliable decisions.

\subsection{Adaptation weakens static evaluation}

AI reviewers are commonly evaluated on fixed collections of manuscripts, where model judgments are compared with historical reviews, scores, or benchmark labels. The manipulation literature shows why these evaluations can become less informative after deployment. Once authors can observe or repeatedly query an AI-mediated evaluator, they can adapt instructions, wording, presentation, or other evaluator-facing signals in ways that change the resulting judgment \citep{zhou2025give,yang2026no,li2026gaming}.

Evaluation should therefore test not only performance on a fixed distribution, but also whether judgments remain reliable under adaptation. Relevant questions include whether semantically similar papers receive consistent evaluations, whether important scientific weaknesses remain influential after presentation changes, and whether robustness persists when participants receive feedback and can change their strategies \citep{dycke2026automatic,jiang2026badscientist}. The central issue is whether an evaluator continues to distinguish stronger scientific work from more effectively optimized presentation once its behavior becomes part of the environment.

\subsection{Safeguards redistribute risk and burden}

The evidence also shows that defensive measures rarely remove risk without changing where it appears. Detection and verification can make particular forms of AI use or manipulation more costly, but they can also motivate evasion. More aggressive enforcement can introduce false positives, while AI-assisted evaluation can create additional verification work, confidentiality concerns, and unequal effects across participants \citep{liang2023gpt,saha2026policies,boonsuk2026editor,vasu2026justice,chen2025envisioning}.

Safeguards should therefore be evaluated by more than their ability to detect or block a targeted behavior. A useful intervention must also account for the human work needed to verify its outputs, the errors it introduces, the participants who bear those errors, and how behavior changes after the intervention is deployed. This is especially important because sophisticated participants may be able to adapt to a control while ordinary users continue to bear its false-positive or compliance costs.

\subsection{Research agenda}

These findings point to a small number of empirical questions that remain unresolved across the current literature. Rather than requiring another set of isolated capability benchmarks, they require studies that directly test the response relations represented in the taxonomy. The four priorities below follow the open links identified by that mapping.

\paragraph{Production scaling and evaluation pressure.}
Existing studies associate AI adoption with greater scholarly output, faster production cycles, and growing submission volume, but the size and generality of the resulting evaluation burden remain uncertain. Multi-venue longitudinal studies with stronger measures of AI use could examine how changes in production affect submission volume, reviewer demand, desk-screening effort, review completion, and the distribution of evaluator attention.

\paragraph{AI-mediated evaluation and author adaptation.}
Controlled studies show that AI reviewers can be manipulated, but there is much less evidence about how authors change their behavior when they know that AI is part of a real evaluation process. Future studies could compare human-only and AI-mediated evaluation settings and measure changes in manuscript presentation, revision behavior, evaluator queries, scientific content, and review outcomes.

\paragraph{Defense and post-policy evasion.}
The literature demonstrates many mechanisms for circumventing detectors and other safeguards, but rarely observes the full sequence after a real policy intervention. Longitudinal studies should track whether a specific venue defense changes participant behavior, which evasion strategies emerge, what false positives or additional burdens result, and how the venue subsequently adjusts its response.

\paragraph{Scholarly records and successor systems.}
Long-horizon feedback remains the least directly traced part of the process. Future work should preserve versioned provenance across scholarly corpora and AI systems so that researchers can determine whether a particular paper, review, correction, or decision entered a later retrieval or training pipeline and materially changed subsequent generation or evaluation. Such studies would distinguish documented reuse from a genuinely closed feedback loop.

These directions shift the research agenda from evaluating isolated AI components toward studying the response relations that connect production, evaluation, manipulation, defense, evasion, and long-horizon feedback. Progress will depend especially on evidence that follows these interactions in real scholarly settings and over time.

\section{Limitations}
\label{sec:limitations}

This survey covers a rapidly changing area, and many relevant studies remain available only as preprints. Our structured literature search covered work available through July 1, 2026. We then conducted a targeted update during the survey development. Because this update did not rerun every search query, coverage of the newest literature may be uneven across taxonomy categories. 

The evidence base also includes surveys, position papers, policy documents, comments, replies, and conceptual analyses. Surveys and conceptual sources are used primarily to characterize terminology, proposals, and ongoing debates, while official policy documents provide evidence of announced institutional responses. Neither type is treated as evidence of real-world prevalence, causal effects, or policy effectiveness unless supported by corresponding empirical data.

Real-world evidence is also uneven across the six dynamics. Large-scale deployments and institutional records are concentrated in a relatively small number of AI and computer-science conferences, OpenReview-based settings, and selected journals, while many attack, defense, and long-horizon feedback studies remain controlled experiments or early demonstrations. Findings from these settings may therefore not transfer directly to disciplines with different review structures, publication timelines, or norms of AI use.

Finally, the paper-review arms-race framing emphasizes adaptation and counter-adaptation between scholarly actors. It is intended as a lens for organizing these interactions rather than as a claim that all AI-assisted research, reviewing, revision, or institutional change is adversarial. Many uses of AI can improve access, efficiency, communication, and scientific quality, and the relevance of the framing depends on whether one actor's behavior meaningfully changes the incentives or responses of another.

\section{Conclusion}
\label{sec:conclusion}

This survey presents AI-mediated scholarly publishing as a connected process rather than a set of isolated tools. We organize the literature around six linked dynamics spanning production scaling, evaluation automation, evaluation manipulation, defense mechanisms and policy responses, evasion and side effects, and long-horizon ecosystem feedback. Together, these dynamics show how changes in one part of the publishing process can reshape the incentives, constraints, and behavior of others. As AI becomes more deeply embedded in research and review, the central challenge is therefore not only whether individual systems perform well, but how they interact and adapt within the broader scholarly ecosystem. Future work should increasingly evaluate these relationships in real settings and over time.

\clearpage
\bibliographystyle{plainnat}
\bibliography{main}

\clearpage
\appendix
\section{Supplemental details}
\label{app:evidence-tables}

\setcounter{table}{0}
\renewcommand{\thetable}{\thesection.\arabic{table}}
\renewcommand{\theHtable}{\thesection.\arabic{table}}
\setlength{\LTcapwidth}{\textwidth}
\makeatletter
\let\AppendixLTMakeCaption\LT@makecaption
\renewcommand{\LT@makecaption}[3]{%
  \AppendixLTMakeCaption{\small#1}{#2}{#3}}
\makeatother

This appendix consolidates the resources and studies discussed throughout the
survey. Table~\ref{tab:data-resources} summarizes datasets, benchmarks, and
deployments. Table~\ref{tab:evidence-map} organizes the surveyed studies and
systems using the taxonomy in Figure~\ref{fig:detailed-taxonomy}.

\subsection{Datasets, benchmarks, and deployments}

\begingroup
\scriptsize
\setlength{\tabcolsep}{2.7pt}
\renewcommand{\arraystretch}{1.04}
\begin{longtable}{@{}
>{\raggedright\arraybackslash}p{0.22\textwidth}
>{\raggedright\arraybackslash}p{0.14\textwidth}
>{\raggedright\arraybackslash}p{0.36\textwidth}
>{\raggedright\arraybackslash}p{\dimexpr0.28\textwidth-6\tabcolsep\relax}@{}}
\caption{Datasets, benchmarks, and deployments discussed in the survey,
organized by resource type, scale, and primary use.}
\label{tab:data-resources}\\
\toprule
Resource & Evidence type & Reported scale & Primary use \\
\midrule
\endfirsthead
\multicolumn{4}{@{}l}{\scriptsize\itshape Table \thetable\ continued from the previous page}\\
\toprule
Resource & Evidence type & Reported scale & Primary use \\
\midrule
\endhead
\midrule
\multicolumn{4}{r@{}}{\scriptsize Continued on the next page}\\
\endfoot
\bottomrule
\endlastfoot

\multicolumn{4}{@{}l}{\textbf{Scholarly records and training corpora}}\\
PeerRead \citep{kang2018dataset} & Corpus & 14,784 papers and 10,770 reviews & Review and decision modeling \\
MReD \citep{shen2022mred} & Annotated corpus & 7,089 meta-reviews, 45K sentences, and 9 labels & Controllable meta-review generation \\
OpenReviewer \citep{idahl2025openreviewer} & Training corpus & 36K papers, 141K collected reviews, and about 79K retained & Scientific review generation \\
DeepReview-13K \citep{zhu2025deepreview} & Training set and benchmark & 13,378 valid samples and a 1.2K-sample test set & Structured review reasoning \\
RMR-75K / RBTAct \citep{wu2026rbtact} & Aligned corpus & 75,542 review-rebuttal mappings across 4,825 papers & Actionable feedback generation \\
GoodPoint \citep{mun2026goodpoint} & Longitudinal corpus & 18,936 ICLR papers, 2020--2026 & Feedback validity and actionability \\
\midrule

\multicolumn{4}{@{}l}{\textbf{Evaluation benchmarks and deployments}}\\
ReviewEval \citep{garg2025revieweval} & Benchmark & 120 papers from NeurIPS, ICLR, and UAI & Five-dimensional review evaluation \\
PRISM \citep{loc2026prism} & Benchmark & 1,000 papers across 5 venue-years, with 5 AI reviewers plus humans & Multi-dimensional reviewer evaluation \\
CLAIMCHECK \citep{ou2025claimcheck} & Expert-annotated benchmark & 41 papers, 60 reviews, and 154 target claims & Claim-grounded critique \\
Soundness counterfactuals \citep{dycke2026automatic} & Counterfactual benchmark & 133 papers and 931 variants (391 critical, 540 neutral) & Faulty-reasoning sensitivity \\
Full-manuscript pre-review \citep{liang2024can} & Retrospective and prospective study & 3,096 Nature-family papers, 1,709 ICLR papers, and 308 researchers & Author-facing feedback \\
ICLR feedback RCT \citep{thakkar2026large} & Randomized deployment & 44,831 reviews, 11,553 papers, and 18,946 feedback recipients & Reviewer assistance \\
AAAI-26 AI Review \citep{biswas2026ai} & Official deployment & 22,977 papers and 5,834 survey responses & Auxiliary AI review \\
FARS \citep{tang2026fars} & Public deployment & 166 complete papers across 67 topics and 282 reviews of 140 papers & End-to-end production and evaluation \\
\midrule

\multicolumn{4}{@{}l}{\textbf{Manipulation, defense, and evasion test beds}}\\
Hidden-prompt test bed \citep{zhu2025your} & Controlled benchmark & 1,441 ICLR and NeurIPS papers & Reviewer manipulation \\
ARO \citep{torrielli2026exploiting} & Controlled benchmark & 5,600 experiments with ChatGPT and Gemini & Cross-role prompt injection \\
PaperGuard \citep{zhao2026does} & Multimodal benchmark & 1,136 papers and 17 in-the-wild attacks with source & Text and figure attacks and defenses \\
SafeReview \citep{xin2026safereview} & Adaptive benchmark & 500 NeurIPS 2024 training papers and 1,286 ICLR 2024 test papers & Co-evolutionary defense \\
Random-citation probe \citep{rao2025detecting} & Randomized audit & Multiple LLMs and review datasets with a test of more than 10K reviews & Outsourced-review detection \\
IntraGuard \citep{ma2026shattering} & Controlled benchmark & 17,844 cases across 7 chatbot settings and 12 venues & Hidden defensive probes \\
Mixed-review attribution \citep{saha2026policies} & Detection benchmark & 50,156 reviews and 5 detectors & Policy enforceability \\
RAID \citep{dugan2024raid} & Robustness benchmark & More than 6M generations from 11 generators, covering 8 domains, 11 attacks, and 12 detectors & Detector evasion \\
\midrule

\multicolumn{4}{@{}l}{\textbf{Long-horizon reuse and impact records}}\\
SPARK \citep{sanyal2025spark} & Training corpus and system & 600K OpenReview reviews with more than 10,000 ideas filtered & Review-informed idea filtering \\
ReviewGuard \citep{rasool2026reviewguard} & Longitudinal benchmark & 20,861 papers, including 2,365 rejected-then-published papers & Citation-aligned evaluation \\
Dolma scholarly component \citep{soldaini2024dolma} & Pretraining corpus & 38.8M Semantic Scholar documents and 70B LLaMA tokens & Scholarly-record ingestion \\
OpenScholar DataStore \citep{asai2026synthesizing} & Retrieval corpus & 45M open-access papers & Citation-backed synthesis \\
Common Pile \citep{kandpal2026common} & Pretraining corpus & 8TB from 30 sources & Licensed scholarly-text reuse \\

\end{longtable}
\endgroup
\clearpage

\subsection{Taxonomy index}

\begingroup
\scriptsize
\setlength{\tabcolsep}{2.7pt}
\renewcommand{\arraystretch}{1.05}
\begin{longtable}{@{}
>{\raggedright\arraybackslash}p{0.16\textwidth}
>{\raggedright\arraybackslash}p{0.21\textwidth}
>{\raggedright\arraybackslash}p{\dimexpr0.63\textwidth-4\tabcolsep\relax}@{}}
\caption{Studies and systems organized by the mechanisms in the taxonomy shown
in Figure~\ref{fig:detailed-taxonomy}.}
\label{tab:evidence-map}\\
\toprule
Dynamic & Mechanism & Studies and systems \\
\midrule
\endfirsthead

\multicolumn{3}{@{}l}{\scriptsize\itshape Table \thetable\ continued from the previous page}\\
\toprule
Dynamic & Mechanism & Studies and systems \\
\midrule
\endhead

\midrule
\multicolumn{3}{r@{}}{\scriptsize Continued on the next page}\\
\endfoot

\bottomrule
\endlastfoot

Production scaling
& AI-assisted and agentic production
& ResearchAgent \citep{baek2025researchagent}, The AI Scientist \citep{lu2024ai}, AI Scientist-v2 \citep{yamada2025ai}, Agent Laboratory \citep{schmidgall2025agent}, FARS \citep{tang2026fars}, PaperOrchestra \citep{song2026paperorchestra}, PaperOrchestrator \citep{yuan2026paperorchestrator} \\

& Faster output and iteration
& Corpus-level LLM estimator \citep{liang2025quantifying}, excess-vocabulary audit \citep{kobak2025delving}, LLM-adoption panel \citep{kusumegi2025scientific}, preprint interrupted time series \citep{qi2025does}, matched-author panel \citep{filimonovic2025can}, Organization Science panel \citep{gartenberg2026more}, ICLR submission series \citep{knight2026flooded} \\

& Post-submission scaling and review pressure
& CycleResearcher \citep{weng2025cycleresearcher}, DRPG \citep{han2026drpg}, Paper2Rebuttal \citep{ma2026paper2rebuttal}, REspGen \citep{ruan2026author}, XtraGPT \citep{chen2026xtragpt} \\
\midrule
Evaluation automation
& Author-facing feedback
& Full-manuscript pre-review \citep{liang2024can}, Focused Feedback Generator \citep{chamoun2024automated}, Checklist Assistant \citep{goldberg2024usefulness}, drafting-feedback study \citep{wu2026drafting}, BOSC pre-review \citep{ohta2026ai} \\

& Reviewer assistance and official review
& ICLR Review Feedback Assistant \citep{thakkar2026large}, AAAI-26 AI Review \citep{biswas2026ai}, OpenReviewer \citep{idahl2025openreviewer}, MReD \citep{shen2022mred} \\

& Grounding, auditability, and reliability
& ReviewAgents \citep{gao2025reviewagents}, PeerCheck \citep{chen2026peercheck}, ReviewGrounder \citep{li2026reviewgrounder}, ScholarPeer \citep{goyal2026scholarpeer}, EGTR-Review \citep{qiu2026egtr}, DeepReviewer 2.0 \citep{weng2026deepreviewer}, PRISM \citep{loc2026prism}, CLAIMCHECK \citep{ou2025claimcheck} \\
\midrule
Evaluation manipulation
& Concealed instructions
& White/microscopic-text injection \citep{lin2026hidden}, positive-review injection \citep{zhou2025give}, invisible-text injection \citep{choi2026invisible}, multilingual hidden injection \citep{theocharopoulos2025multilingual}, malicious-font injection \citep{xiong2025invisible}, indirect prompt injection \citep{torrielli2026exploiting} \\

& Presentation and multimodal steering
& Adversarial Repackaging \citep{yang2026no}, abstract-rephrasing attack \citep{li2026gaming}, paraphrasing attack \citep{kaneko2026paraphrasing}, Figure-only attack \citep{zhao2026does}, textual adversarial perturbations \citep{lin2025breaking} \\

& Adaptive evaluator optimization
& LLM-REVal \citep{li2025llm}, Review Arcade \citep{hatzel2026review}, aspect-guided perturbation \citep{li2025llms}, optimization-based prompt injection \citep{shi2024optimization}, universal adversarial prompt \citep{raina2024llm}, surrogate-feedback refinement \citep{zhou2025give} \\
\midrule
Defense mechanisms and policy responses
& Detection and targeted verification
& Quis \citep{kumar2024quis}, Sem-Detect \citep{duarte2026sem}, MixRevDetect \citep{kumar2025mixrevdetect}, CoCoNUTS \citep{chen2026coconuts}, random-citation watermark \citep{rao2025detecting}, IntraGuard \citep{ma2026shattering}, topic-based watermarking \citep{nemecek2025feasibility} \\

& Evaluator and artifact protection
& SafeReview \citep{xin2026safereview}, PaperGuard \citep{zhao2026does}, TAB-AUDIT \citep{huang2026tab}, SciFigDetect \citep{hu2026scifigdetect}, detection prompt \citep{zhou2025give} \\
\midrule
Evasion and side effects
& Post-generation evasion
& RAID \citep{dugan2024raid}, DIPPER \citep{krishna2023paraphrasing}, recursive paraphrasing \citep{sadasivan2025can}, RAFT \citep{wang2024raft}, CoPA \citep{fang2025your} \\

& Generation-time and route adaptation
& Adversarial humanization \citep{zhou2024humanizing}, proxy attack \citep{wang2025humanizing}, GradEscape \citep{meng2025gradescape}, style-shifting attack \citep{pedrotti2025stress}, MASH \citep{gu2026mash}, constrained policy optimization \citep{wang2026detector} \\

& Redistributed error, labor, and risk
& Non-native-writer bias \citep{liang2023gpt}, accuracy-bias trade-off \citep{pratama2025accuracy}, authorship misclassification \citep{boonsuk2026editor}, AI-use stigmatization \citep{ali2026stigmatization}, reviewer-attribute bias \citep{vasu2026justice}, verification workload \citep{chen2025envisioning} \\
\midrule
Long-horizon ecosystem feedback
& Record ingress and persistence
& Published-article diffusion audit \citep{siler2026diffusion}, knowledge-update phrase audit \citep{strzelecki2025my}, Google Scholar paper audit \citep{haider2024gpt}, undeclared-AI audit \citep{yao2026ai}, false-authorship case audit \citep{spinellis2025false}, fabricated-citation audit \citep{topaz2026fabricated}, citation-propagation tracing \citep{camp2025citation} \\

& Reuse in future AI systems
& Dolma/OLMo \citep{soldaini2024dolma}, OpenScholar \citep{asai2026synthesizing}, Common Pile \citep{kandpal2026common}, RBTAct \citep{wu2026rbtact}, GoodPoint \citep{mun2026goodpoint}, SPARK \citep{sanyal2025spark}, ReviewGuard \citep{rasool2026reviewguard} \\

& Propagation and the open causal loop
& Knowledge-graph poisoning \citep{yang2024poisoning}, PoisonedRAG \citep{zou2025poisonedrag}, medical-model data poisoning \citep{alber2025medical}, retrieval-pollution test \citep{yu2026retrieval}, recursive model collapse \citep{shumailov2024ai}, self-consuming loop \citep{alemohammad2024self}, diversity-collapse test \citep{guo2024curious}, Artificial Hivemind \citep{jiang2026artificial} \\

\end{longtable}
\endgroup
\FloatBarrier


\end{document}